\documentclass{article}
\usepackage{newunicodechar}
\newunicodechar{−}{\ensuremath{-}}
\usepackage{mathtools}
\usepackage{amsmath,amssymb,amsfonts}
\usepackage{algorithmic}
\usepackage{graphicx,color} % Required for inserting images
\usepackage{booktabs}
\usepackage{tabularx}
\usepackage{wrapfig}
\usepackage{textcomp}
\usepackage{csvsimple}
\usepackage{hyperref}
\usepackage{balance}
\usepackage[utf8]{inputenc}
\usepackage{textgreek}
\usepackage{nomencl}
\usepackage{float}
\usepackage{subcaption}
\usepackage[sorting=none,backend=biber,style=numeric]{biblatex}
\makenomenclature
\usepackage[table]{xcolor}
\usepackage{bm} % for bold math symbols

\title{\textbf{Hybrid Classical-Quantum(HCQ) Alzheimer's Classification via Supervised $\bm{\beta}$-VAE and Quantum Kernels}}

\author{
    Tia Tiwari$^{*1}$ \quad
    Vamshi Krishna Kancharla$^{\dagger 2}$ \quad
    Neelam Sinha$^{\ddagger 1}$ \\[0.5em]
    $^1$ Centre for Brain Research, Indian Institute of Science \\
    $^2$ Vision and AI Lab (VAL), Indian Institute of Science
}

\begin{document}

\maketitle

\begin{abstract}
This paper presents a two-stage Hybrid Classical- Quantum (HCQ) pipeline for binary Alzheimer's disease (AD) classification from 3D T1-weighted structural MRI volumes, where the classical and quantum components are designed to complement each other rather than operate independently. A supervised 3D $\beta$-variational autoencoder (VAE) is trained end-to-end under voxel-wise reconstruction, KL-divergence, and focal classification losses that compress each 3D MRI volume (resized from $152 \times 184 \times 152$ to $96 \times 96 \times 96$) into a 64-dimensional latent code. Partial Least Squares (PLS) regression selects the six components in the latent code that best separate Alzheimer's Disease (AD) from cognitively normal (CN) subjects and rescales them into rotation angles, which are encoded onto a six-qubit register using the ZZ quantum feature map to give us the respective quantum states. An input for a precomputed-kernel Support Vector Machine (SVM) is an $N \times N$ Gram matrix ($N = 308$), created by calculating the overlap between every pair of quantum states. The novelty of this work lies in the fact that the quantum kernel operates directly on disease-aware features learned end-to-end by a supervised autoencoder, rather than on pre-extracted inputs. On 308 ADNI-1 subjects, consisting of 137 AD and 171 CN subjects, the baseline achieved 67.2\% accuracy and 0.759 AUC, while the stability-enhanced variant reached 72.1\% accuracy and 0.799 AUC with cross-fold variance halved. 3D Grad-CAM further helped validate our model's focus on brain regions linked to Alzheimer's. The HCQ pipeline could serve as a general-purpose framework for diagnostic classification across biomedical imaging domains that present similar challenges for classical approaches.
\end{abstract}

\section*{Keywords}
3D Structural MRI volume; Alzheimer's disease; Hybrid classical-quantum model; $\bm{\beta}$-Variational Autoencoder; Quantum Kernel.

\section{Introduction}
\label{sec:introduction}

Alzheimer's disease(AD) is a common neurodegenerative disease. The progression of this disease produces notable structural changes in the brain, which is atrophy of the hippocampus, degeneration of medial temporal regions and progressive thinning of the cerebral cortex~\cite{frisoni2010clinical}. These changes are captured non-invasively using structural Magnetic Resonance Imaging (sMRI), which is a foundational tool for not just for clinical assessment but even computational research in AD diagnosis ~\cite{frisoni2010clinical}.The computational problem built on this imaging foundation is treated as a binary classification task, where a given T1-weighted MRI is used to classify subjects as Alzheimer's Disease(AD) or Cognitively Normal(CN). The data used in this study was obtained from Alzheimer's Disease Neuroimaging Initiative-1 (ADNI-1)~\cite{petersen2010alzheimer}, a standardized benchmark widely adopted for research based on AD classification.

Despite considerable efforts in multiple research communities, the task to classify remains constrained due to insufficient data availability~\cite{alsubaie2024alzheimer}. MRI data acquisition and confirming a clinical diagnosis is time-consuming and expensive, resulting in cohort sizes rarely exceeding a few hundred subjects. This raises a genuine question: Should a model try to use every detail in a 3D MRI scan and risk overfitting on limited data or switch to a simpler model that inevitably discards diagnostically relevant information? The field of quantum machine learning (QML) has recently emerged to show promise in this kind of low-data setting. 

\subsection{Related Work}
The research on automatic diagnosis of Alzheimer's disease (AD) through the lens of structural MRI has progressed through a number of methodological rounds, which have traded off model complexity with the quantity of data necessary. In the literature, a general development pattern appears from manual feature engineering, moving to end-to-end deep learning, then to learned compression methods, and recently to hybrid classical-quantum methods.

The early computational methods used handcrafted features and classical classifiers. Anatomical data like hippocampal volume and regional cortical thickness are retrieved through existing image-processing pipelines and classified with shallow learners, like Support Vector Machines (SVMs). The biomarkers themselves are clinically validated~\cite{frisoni2010clinical}, classifiers remain interpretable, and the pipeline as a whole works reliably even on small cohorts. Despite its advantages, the methodology is fundamentally limited by the choice of measurements of the practitioner. Any pathological signal residing outside the explicitly quantified regions of interest is unavailable to the model. Performance thus plateaus no matter the size or quality of the dataset, suggesting that improving performance will take more work to discover disease-relevant structure instead of building around predefined features.

This motivated the onset of  three-dimensional convolutional neural networks (CNNs), such as recent attention-augmented ones~\cite{zhang2023attention}, that learn discriminative features directly from raw volumetric MRI inputs. On large datasets, these models achieve state-of-the-art accuracy. But the high parametric capacity that enables such models to be expressive also results in a demand for data, and clinical AD cohorts seldom exceed a few hundred subjects. Systematic reviews of the field have repeatedly shown that reported deep-learning results on small AD datasets often fail to reproduce when evaluated under rigorous, leakage-free validation protocols~\cite{alsubaie2024alzheimer,wen2020convolutional,liu2021development}. The expressiveness gained if handcrafted features are abandoned is thus at least partially offset by the data efficiency it sacrifices, demonstrating that scale alone cannot resolve the underlying trade-off.

A third line of work deals with this using compress-then-classify pipelines, where an autoencoder first reduces each 3D MRI volume to a compact latent representation, and a different model subsequently performs the classification~\cite{bit2024mri}. This decoupling significantly reduces the number of parameters used by the downstream classifier and thus mitigates overfitting. However, traditional autoencoders are trained only to reconstruct the input, with no function related to a diagnostic label. Thus, the latent space learned encodes general anatomical variation rather than disease-discriminative structure, and the pathologically relevant signal might be lost during compression. The core trade-off remains: what is gained in data efficiency is paid for by the loss of disease-specific signal.

Most recently, quantum machine learning (QML) has emerged as a fundamentally different paradigm for medical-image analysis~\cite{schuld2019quantum}. Quantum kernels embed classical inputs into the exponentially large Hilbert space of an $n$-qubit system~\cite{schuld2019quantum,havlivcek2019supervised}, enabling the representation of similarity relationships that classical kernels cannot capture. Indeed, supervised quantum models of this form are provably kernel methods~\cite{schuld2021supervised}, and recent theoretical work has shown that they can generalize from remarkably few training samples~\cite{caro2022generalization}an attribute of particular relevance in the low-data regime that characterizes clinical neuroimaging. This motivation has produced a growing body of hybrid classical-quantum (HCQ) work applied to AD diagnosis and broader neuroimaging tasks. Reported approaches include deep ensembles coupled with quantum classifiers~\cite{jenber2024deep}, general hybrid quantum deep-learning architectures for medical images~\cite{mazher2024hybrid}, dedicated quantum screening models for AD~\cite{cappiello2025quantum}, pipelines combining classical autoencoders with quantum SVMs on high-dimensional neuroimaging data~\cite{chen2025compressedmediq}, and lightweight hybrid classical-quantum CNNs for 3D structural MRI~\cite{islam2025cq,barua2025hybrid}. Two limitations consistently characterize this body of work. First, the qubit constraints of current Noisy Intermediate-Scale Quantum (NISQ) devices and simulators~\cite{preskill2018quantum} preclude direct ingestion of 3D MRI volumes, which contain on the order of $10^{6}$ voxels , requiring aggressive dimensionality reduction prior to any quantum operation. Second, and more substantively, the quantum components in these pipelines operate predominantly on pre-extracted, handcrafted, or unsupervised features rather than on representations trained end-to-end to be discriminative for the diagnostic task. Furthermore, fidelity-based quantum kernels are known to suffer from exponential concentration as qubit count grows~\cite{thanasilp2024exponential,schnabel2025quantum}, and recent benchmarking studies suggest that trainable quantum embedding kernels~\cite{hubregtsen2022training} may be needed to realize practical advantages over classical alternatives.

Taken together, the surveyed paradigms address one aspect of the data-expressiveness trade-off without resolving it. The handcrafted pipelines preserve interpretability and data efficiency but cannot discover unmeasured signal; end-to-end CNNs are expressive but data-hungry; unsupervised autoencoders are data-efficient but disease-agnostic; and existing hybrid quantum methods, while introducing a complementary paradigm, have not yet been evaluated on representations that are simultaneously learned, compact, and disease-aware. The methodological gap is therefore precise: a hybrid pipeline in which the classical stage produces a compact, supervised, disease-discriminative representation, and the quantum stage operates directly on that representation rather than on features prepared in advance. The framework introduced in the following subsections is designed to close this gap.

\subsection{Motivation}

The analysis made in the earlier section hints at a methodological gap that the existing approach does not cover. Handcrafted feature pipelines are efficient but have limits in data usage due to manually injected measurements.
Deep CNNs are powerful but data-intensive; unsupervised autoencoders reduce the dimensionality but fail to retain disease-specific information; and current hybrid quantum models have not been tested on representations that are simultaneously learned, compact, and disease-aware. We aim to address this gap in this work by providing a representation of all three characteristics and coupling it to a quantum kernel classifier. The quantum component is placed where its theoretical strengths are most relevant, in a small-data setting where the inputs are compact, learned features, and where the exponential dimensionality of the quantum feature space and its ability to capture non-classical similarity patterns are most likely to provide value.

\subsection{Proposed Framework}

\begin{figure}[H]
   \centering
    \includegraphics[width=\textwidth]{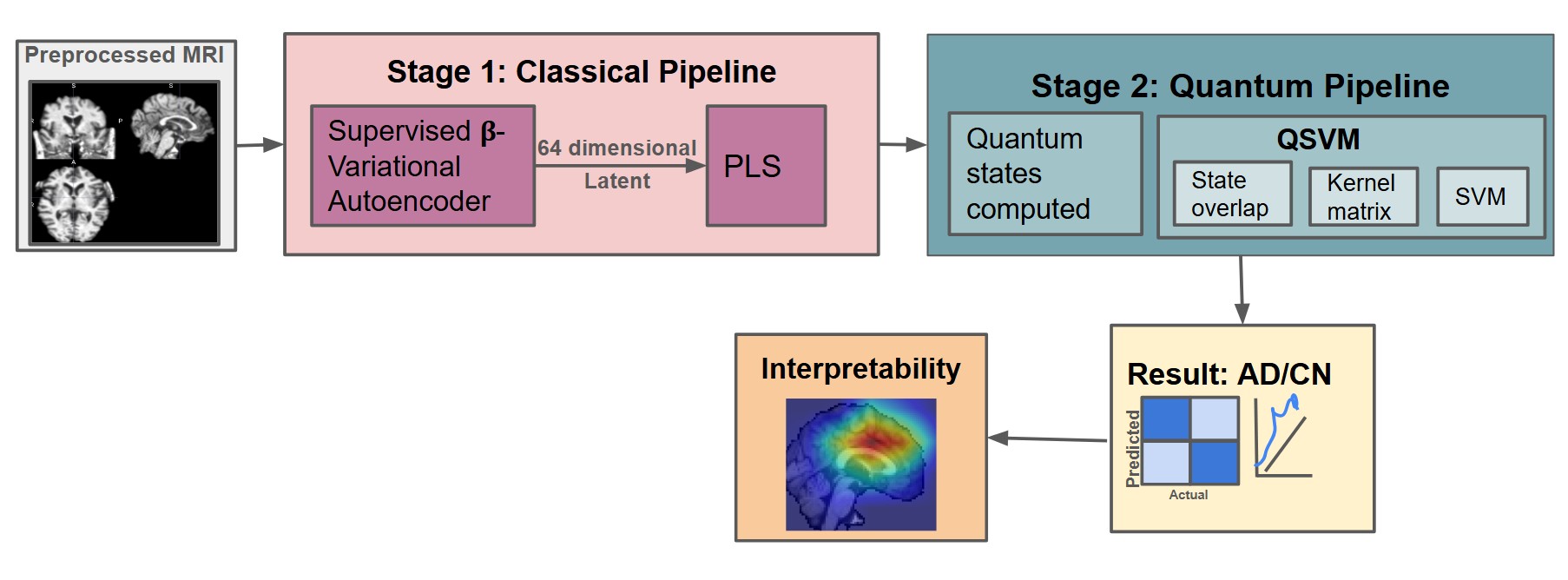}
   \caption{Overview of the proposed two-stage hybrid classical-quantum (HCQ) pipeline. The classical stage compresses each 3D MRI volume into a compact latent representation through a supervised $\beta$-VAE and a label-aware projection. The quantum stage encodes this representation through a quantum feature map and classifies subjects via the resulting quantum-kernel SVM. 3D Grad-CAM is used to perform interpretability.}
   \label{fig:overview}
\end{figure}

We propose a two-stage hybrid classical-quantum (HCQ) pipeline for our classification problem, which is designed to close the methodological gap identified above. The framework is further organized as two complementary major stages. The classical stage learns a compact, disease-aware representation of each MRI volume that addresses the limitation of unsupervised autoencoders. The quantum stage then operates directly on this learned representation, addressing the limitation of prior hybrid approaches that relied on pre-extracted or handcrafted inputs. The complete pipeline is illustrated in Figure~\ref{fig:overview}.

In the classical stage, a supervised $\beta$-Variational Autoencoder(VAE) compresses each preprocessed 3D MRI volume into a low-dimensional latent code trained jointly for reconstruction, latent regularization, and AD-versus-CN classification; a label-aware projection then reduces this code to a small set of diagnostically informative features. These features then enter the quantum stage, where a feature-map circuit encodes them onto a qubit register in a high-dimensional Hilbert space, pairwise quantum-state overlaps form a similarity matrix, and a Support Vector Machine trained on that matrix produces the final prediction. A parallel 3D Grad-CAM branch verifies that predictions correspond to anatomically meaningful brain regions. The complete pipeline is shown in Figure~\ref{fig:overview}.

The architecture deliberately positions each component where its theoretical advantage is most likely to translate into practical benefit. We use classical deep learning where representation discovery is needed and a quantum kernel where small data similarity-based classification is best supported by theory.

\section{Dataset}

The data used in this study was obtained from a subset of ADNI-1 cohort. Since the present work addresses binary AD-versus-CN classification, MCI subjects were excluded, and the remaining AD and CN scans were subjected to quality control, yielding a final subset of 308 subjects with 137 AD and 171 CN , with an approximately balanced sex distribution, summarized in Table~\ref{tab:cohort}. The mild class imbalance toward CN is addressed during training (Section~\ref{sec:methodology}) rather than by discarding subjects. Each subject contributes a single T1-weighted MRI volume.
\begin{table}[ht]
    \centering
    \caption{Subject distribution in the ADNI-1 cohort.}
    \label{tab:cohort}
    \begin{tabular}{lccc}
        \toprule
        \textbf{Group} & \textbf{Male} & \textbf{Female} & \textbf{Total} \\
        \midrule
        AD    & 70  & 67  & 137 \\
        CN    & 89  & 82  & 171 \\
        \midrule
        \textbf{Total} & 159 & 149 & \textbf{308} \\
        \bottomrule
    \end{tabular}
\end{table}

\newpage
\section{Methodology}
\label{sec:methodology}
This section details the two stages of the proposed pipeline:
classical feature extraction (Stage-1) and quantum kernel 
classification (Stage-2). Together, these form the baseline 
configuration \textbf{M1}. A stability-enhanced variant 
\textbf{M2}, which modifies only the classical side, is 
presented in Section~\ref{sec:enhanced}.
\subsection{Classical Feature Extraction}
\label{sec:stage1}

\begin{figure}[H]
    \centering
    \includegraphics[width=\textwidth]{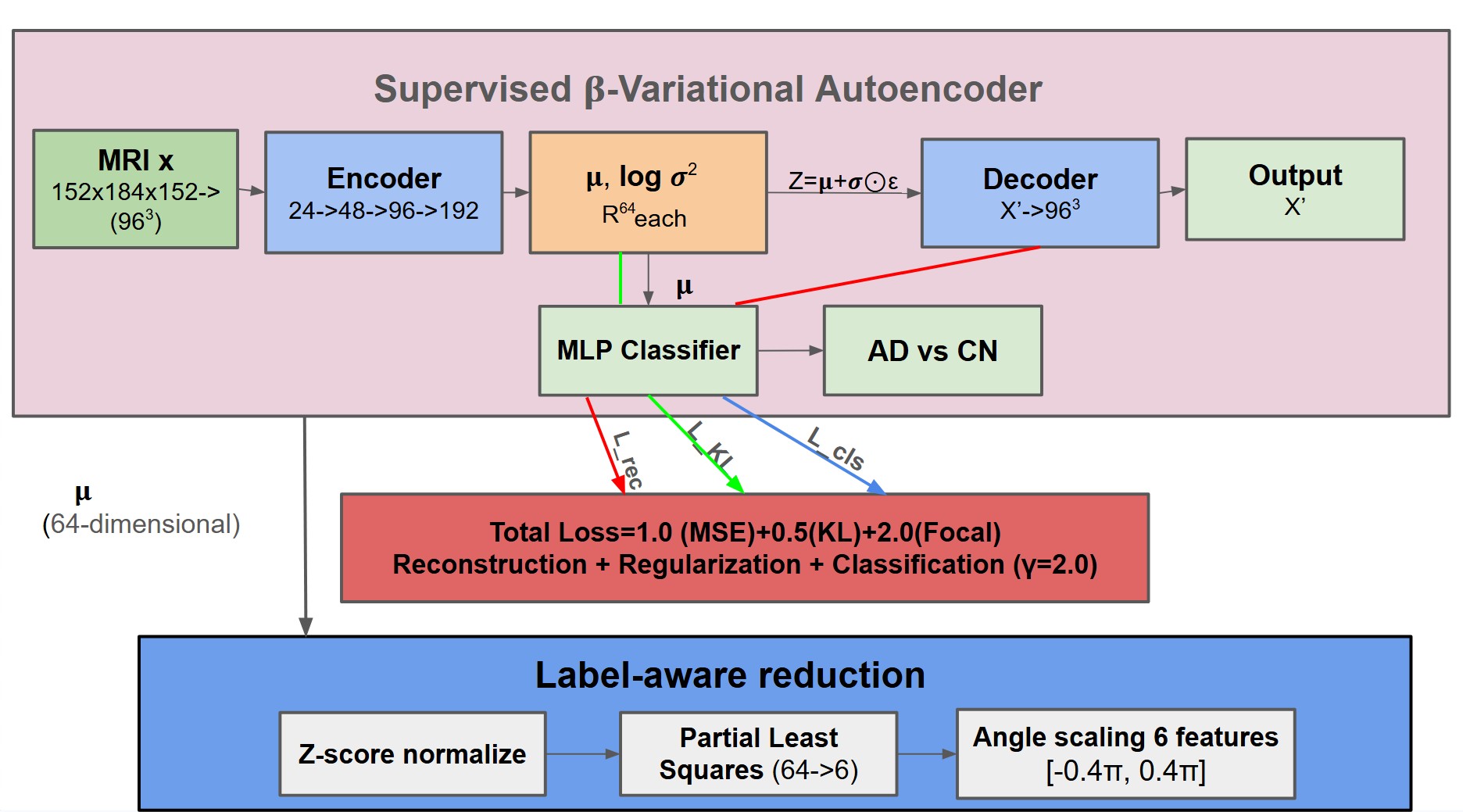}
    \caption{\textbf{Stage-1 architecture}: A 3D encoder compresses each MRI volume into a latent distribution; a decoder reconstructs the volume while a classifier predicts AD versus CN under three joint losses (reconstruction, KL, and focal). PLS then reduces the latent mean μ to six rotation angles in [−0.4π, +0.4π] for the quantum stage.}
    \label{fig:stage1}
\end{figure}

 In Stage 1, a single 3D MRI volume is turned into six rotation angles per subject. A supervised $\beta$-VAE first learns a 64-dimensional  latent code, and PLS reduces it to six components ready for quantum encoding. Figure~\ref{fig:stage1} shows the block-level architecture.
\newpage
 Each subject contributes a T1-weighted MRI volume, originally 
$152 \times 184 \times 152$ voxels after the standard preprocessing 
pipeline. These volumes are resized to a common $96 \times 96 \times 96$ 
grid using trilinear interpolation, which smoothly averages neighboring 
voxels to preserve anatomical detail at the reduced resolution, and 
intensity-normalized within the brain mask. Even at this reduced size, 
each volume contains 884,736 voxels,that is far too many for a quantum circuit to accept directly, which is why we rely on the classical stage to compress these voxels reasonably. A four-block 3D convolutional encoder progressively shrinks the volume while extracting increasingly abstract features. Each block applies a strided convolution (which simultaneously filters and downsamples), instance normalization (which stabilizes training when batch sizes are small), and a LeakyReLU activation (which introduces non-linearity while avoiding dead neurons). The channel width doubles at each block ($24 \to 48 \to 96 \to 192$) while the spatial resolution halves, compressing the $96^3$ input into a compact $6^3$ feature grid.

The flattened $6^3$ grid is passed through two separate linear layers. One produces a 64-dimensional mean vector $\mu \in \mathbb{R}^{64}$, the other a 64-dimensional log-variance vector $\log\sigma^2 \in \mathbb{R}^{64}$. Together, these define a Gaussian distribution $\mathcal{N}(\mu, \sigma^2)$ over possible latent codes for each subject. Instead of mapping each brain to a single fixed point, the encoder maps it to a region in latent space, a cloud of possible codes centered at $\mu$ with spread controlled by $\sigma^2$. We predict $\log\sigma^2$ rather than $\sigma^2$ directly so that the implied variance $\sigma^2 = \exp(\log\sigma^2)$ always remains positive, preventing numerical issues during training. To draw a sample from this distribution during training, the reparameterization trick expresses the sample as
\begin{equation}
    z = \mu + \sigma \odot \varepsilon, \qquad \varepsilon \sim \mathcal{N}(0, I),
    \label{eq:reparam}
\end{equation}
where $\odot$ denotes element-wise multiplication and $\varepsilon$ is drawn from a fixed standard-normal distribution. This separates the randomness (in $\varepsilon$) from the learned parameters ($\mu$ and $\sigma$), allowing the network to be trained end-to-end through standard backpropagation even though sampling itself is not differentiable.

A symmetric decoder made of four transposed-convolution blocks takes the sampled code $z$ and expands it back to a full $96^3$ volume $\hat{x}$-if the reconstruction is faithful, the 64-dimensional code has captured enough anatomical information. In parallel with the decoder, a small multilayer perceptron (MLP) with dropout $0.5$ reads the latent mean $\mu$ and predicts whether the subject has AD or is CN. The classifier reads $\mu$ rather than the sampled $z$ because $\mu$ is the stable center of the distribution, free from the noise introduced by sampling. At inference time, sampling is skipped entirely and all downstream operations use $\mu$ directly. This classifier is what makes the autoencoder \emph{supervised}-a standard VAE would only learn to reconstruct, and its latent space might organize around brain size or head shape rather than disease. By adding the classifier, we force the encoder to arrange the latent space so that AD and CN subjects are separable from the very first training epoch.

The network is trained end-to-end under a composite loss that balances three goals:
\begin{equation}
    \mathcal{L}_{\text{total}} = 1.0\,\mathcal{L}_{\text{MSE}} + 0.5\,\mathcal{L}_{\text{KL}} + 2.0\,\mathcal{L}_{\text{focal}}.
    \label{eq:totalloss}
\end{equation}
The first term, $\mathcal{L}_{\text{MSE}} = \lVert x - \hat{x} \rVert_2^2$, measures how well the decoder rebuilds the original scan and keeps the latent code grounded in real anatomy, without it, the encoder could produce codes that help classification but carry no meaningful anatomical information. The second term, the KL divergence
\begin{equation}
    \mathcal{L}_{\text{KL}} = -\tfrac{1}{2} \sum_{i=1}^{64} \left(1 + \log\sigma_i^2 - \mu_i^2 - \sigma_i^2\right),
    \label{eq:kl}
\end{equation}
measures how far the encoder's distribution $\mathcal{N}(\mu, \sigma^2)$ has drifted from a standard-normal prior $\mathcal{N}(0, I)$. Minimizing this keeps the latent space smooth and continuous, ensuring nearby codes decode to similar brains with no empty gaps. The weight $\beta = 0.5$ is moderate to strong enough to organize the space, gentle enough not to crush reconstruction quality. This smoothness matters downstream because the quantum circuit expects inputs from a well-behaved, evenly distributed range. The third term, the focal loss
\begin{equation}
    \mathcal{L}_{\text{focal}} = -\alpha_t\,(1 - p_t)^{\gamma}\,\log(p_t),
    \label{eq:focal}
\end{equation}
addresses the mild class imbalance (137 AD versus 171 CN) and the difficulty of borderline cases simultaneously. The modulating factor $(1 - p_t)^{\gamma}$ with $\gamma = 2.0$ automatically reduces the loss for confidently classified subjects and concentrates learning on hard, near-boundary cases, while the class-balanced weight $\alpha_t$, set inversely to class frequency, gives the smaller AD group a stronger influence. The weight $\lambda_{\text{cls}} = 2.0$ reflects that classification is the primary goal of the pipeline. Because all three losses are active from the first epoch, the latent space is disease-aware by construction—not as a post-hoc adjustment bolted on after unsupervised pretraining.

The composite loss produces a 64-dimensional code per subject, but a six-qubit quantum circuit cannot accept 64 inputs at practical depth. The classical stage therefore concludes with a supervised dimensionality reduction from 64 to 6. The choice of method matters. Principal component analysis (PCA) would keep the directions of greatest variance, but the highest-variance direction might capture overall brain size rather than disease.  Partial least squares (PLS) regression was used instead, which accounts for the diagnostic label during reduction by finding the six directions that maximize the covariance between the latent codes and the AD/CN labels, ensuring that the values reaching the quantum stage are the most diagnostically informative. The reduction unfolds in three steps: First, the 64-dimensional means $\mu$ are z-score normalized , using a scaler fitted only on the training fold; Second, PLS projects the normalized codes into a 6-dimensional subspace and finally, each component is rescaled into $[-0.4\pi, +0.4\pi]$. This bound is deliberately conservative: rotation angles in a quantum circuit repeat every $2\pi$, so angles too close to $\pm\pi$ could wrap around and make genuinely different subjects look identical to the circuit.

The output of Stage 1 is six rotation angles per subject. The supervised $\beta$-VAE is implemented in PyTorch~\cite{paszke2019pytorch}, and the quantum kernel pipeline follows the PennyLane kernel-training workflow~\cite{pennylane2026demo}.

\subsection{Quantum Kernel Classification}
\label{sec:stage2}

\begin{figure}[H]
    \centering
    \includegraphics[width=\textwidth]{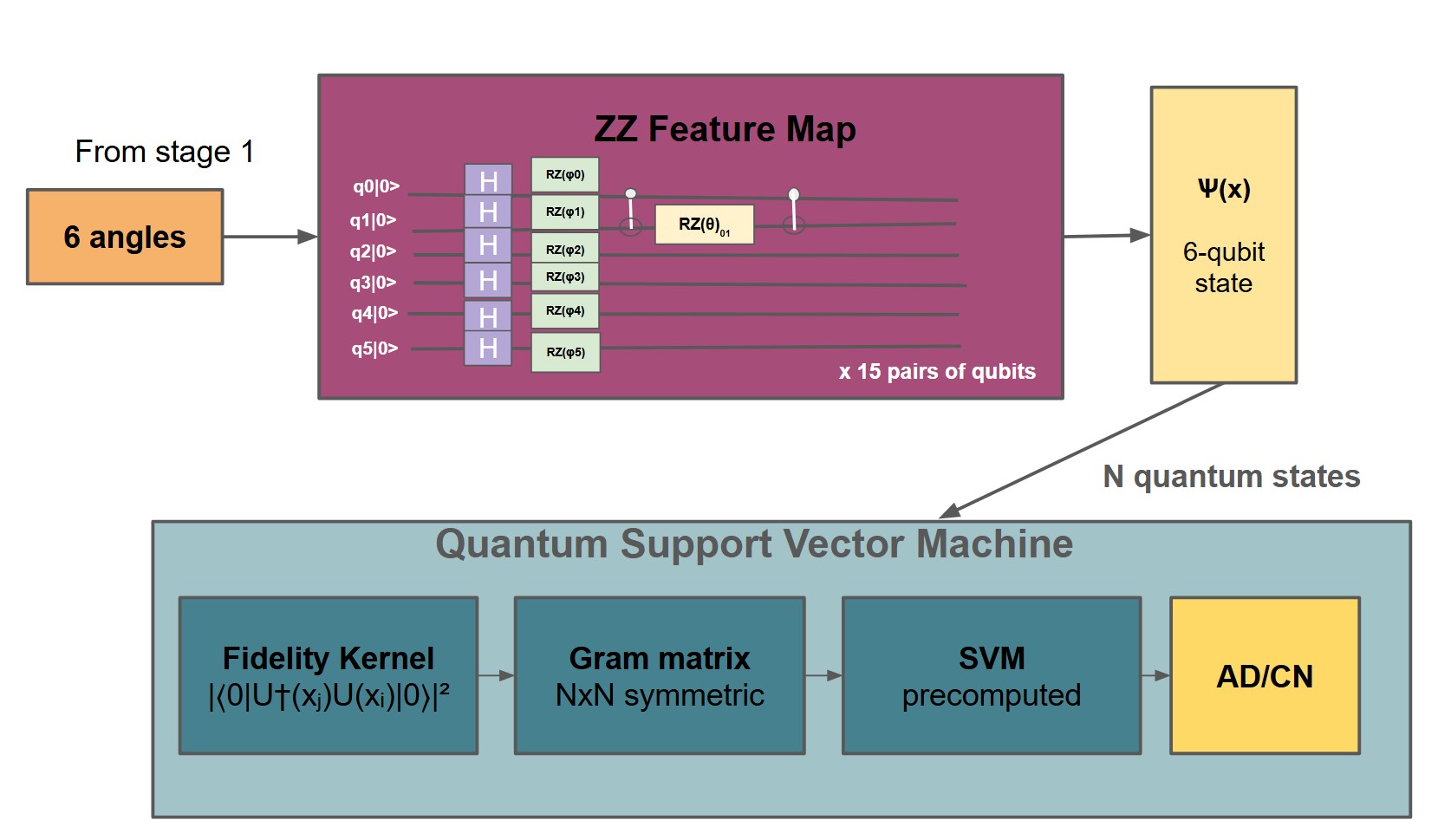}
    \caption{\textbf{Stage-2 architecture.} A ZZ feature map encodes each subject's six angles onto a six-qubit register, embedding the data into a $2^6$-dimensional Hilbert space through superposition and entangling pairwise interactions. The squared overlap between every pair of resulting quantum states fills an $N \times N$ Gram matrix, which a precomputed-kernel SVM uses to produce the final AD-versus-CN prediction.}
    \label{fig:stage2}
\end{figure}

Stage 2 takes the six rotation angles from Stage 1 as input and produces an AD vs CN prediction through three components: A feature-map circuit that encodes the angles into a quantum state, a kernel computation that measures pairwise state overlaps, and an SVM trained on the resulting similarity matrix. Figure~\ref{fig:stage2} shows the block-level architecture.

% ============================================================
% FIGURE PLACEHOLDER --- Stage 2 block diagram
% Source: SVG diagram of ZZ feature map + fidelity kernel + SVM
% ============================================================

The six angles per subject are encoded onto a six-qubit register through a ZZ feature map~\cite{havlivcek2019supervised}, implemented in PennyLane~\cite{bergholm2018pennylane} on the \texttt{lightning.qubit} simulator.The circuit applies a Hadamard to every qubit, writes each angle into its qubit's phase with an $R_Z(\varphi_i)$ rotation, and then applies entangling two a 2 qubit ZZ gates to every pair and each gate realized as $\mathrm{CNOT} \cdot R_Z(2(\pi - \varphi_i)(\pi - \varphi_j)) \cdot \mathrm{CNOT}$. The resulting state $|\varphi(x)\rangle$ encodes both individual feature values and their pairwise interactions.

For every pair of subjects, the fidelity kernel is the squared overlap of their quantum states:
\begin{equation}
    k(x_i, x_j) = \bigl|\langle 0|\, U^{\dagger}(x_j)\, U(x_i)\, |0\rangle\bigr|^2,
    \label{eq:kernel}
\end{equation}
obtained by running one subject's encoding circuit followed by the inverse of the other's and measuring the probability of returning to the all-zeros state. Repeating this for all pairs fills a symmetric $N \times N$ Gram matrix, the entire output of the quantum stage.

The Gram matrix is passed to a Support Vector Machine in precomputed-kernel mode~\cite{schuld2021supervised}, operating only on the quantum-derived similarities. The regularization parameter $C$ is tuned over $\{0.5, 1, 2, 5, 10, 20\}$, and the decision threshold is calibrated on held-out inner-validation data within each cross-validation fold.

\subsection{Stability-Enhanced Variant (M2)}
\label{sec:enhanced}

A practical issue arises when a large 3D neural network is trained on only a few hundred subjects because the outcome is mostly dependent on chance. A different random seed can lead the encoder to organize its latent space differently, and this difference carries through PLS, into the quantum kernel, and ultimately into the final prediction. Two runs of the same pipeline on the same data can therefore produce noticeably different accuracy values, not because the method is flawed, but because the training process is inherently noisy at this sample size. The stability-enhanced variant (M2) is designed to reduce this noise while keeping the quantum architecture entirely unchanged using the same ZZ feature map, same six qubits, same fidelity kernel, and same Gram matrix. It modifies only the classical training and inference through four changes. Instead of training a single $\beta$-VAE per fold, three copies are trained from different random seeds, and only the one with the best inner-validation balanced accuracy is kept, reducing the risk of a poor initialization dominating the result. At inference time, each test subject is passed through the encoder three times with slightly different augmentations (random flips and small additive noise), and the resulting latent codes are averaged to produce a more stable representation that is less sensitive to any single view of the input. Platt scaling fits a simple logistic function to the raw SVM decision scores using held-out validation data, converting them into properly calibrated probabilities—since a raw SVM score of 0.7 does not necessarily mean a 70\% chance of AD, this calibration makes the decision threshold more meaningful.The SVM regularization grid is widened by adding two smaller values ($\{0.1, 0.3\}$) to the baseline grid of $\{0.5, 1, 2, 5, 10, 20\}$, allowing the optimizer to find softer margins that generalize better on limited data. Because none of these modifications touch the quantum architecture, any improvement in accuracy or reduction in cross-fold variance observed in M2 can be attributed specifically to classical training stability rather than a change in the quantum model.

\subsection{Experimental Setup}
\label{sec:setup}

All experiments use five-fold stratified cross-validation over the 308 ADNI-1 subjects, with class proportions preserved in every fold. Table~\ref{tab:setup} summarizes the complete pipeline configuration.

\begin{table}[H]
    \centering
    \caption{Experimental configuration of the HCQ pipeline.}
    \label{tab:setup}
    \small
    \begin{tabular}{p{4.5cm} p{5.5cm}}
        \toprule
        \textbf{Parameter} & \textbf{Value} \\
        \midrule
        \multicolumn{2}{l}{\emph{Data \& Preprocessing}} \\[0.3em]
        Input volume        &  $152 \times 184 \times 152$ \\
        Subjects            & 308 (137 AD / 171 CN) \\
        Augmentation        & Flip ($p{=}0.5$), noise ($\sigma{=}0.10$) \\
        Validation          & 5-fold stratified CV \\
        \midrule
        \multicolumn{2}{l}{\emph{Supervised $\beta$-VAE}} \\[0.3em]
        Encoder / decoder   & 4 blocks, 3D conv / deconv \\
        Channels            & $24 \to 48 \to 96 \to 192$ \\
        Bottleneck          & $6^3$ feature grid \\
        Latent dimension    & 64 \\
        Classifier          & MLP, dropout $0.5$ \\
        Optimizer           & Adam, weight decay $10^{-3}$ \\
        Learning rate       & $10^{-4}$, cosine schedule \\
        Epochs              & 60 \\
        Batch size          & 4 \\
        $\lambda_{\text{rec}}$ & 1.0 \\
        $\beta$ (KL weight) & 0.5 \\
        $\lambda_{\text{cls}}$ & 2.0 \\
        Focal $\gamma$      & 2.0 \\
        \midrule
        \multicolumn{2}{l}{\emph{Dimensionality Reduction}} \\[0.3em]
        Method              & PLS regression \\
        Components          & 6 \\
        Scaling (pre-PLS)   & Z-score normalization \\
        Scaling (post-PLS)  & Min-max to $[-0.4\pi, +0.4\pi]$ \\
        \midrule
        \multicolumn{2}{l}{\emph{Quantum Kernel}} \\[0.3em]
        Simulator           & \texttt{lightning.qubit} \\
        Qubits              & 6 \\
        Feature map         & ZZ \\
        Repetitions         & 1 \\
        \midrule
        \multicolumn{2}{l}{\emph{SVM Classifier}} \\[0.3em]
        Kernel mode         & Precomputed \\
        $C$ grid (M1)       & $\{0.5, 1, 2, 5, 10, 20\}$ \\
        $C$ grid (M2)       & Adds $\{0.1, 0.3\}$ \\
        Threshold           & Calibrated per fold \\
        \midrule
        \multicolumn{2}{l}{\emph{M2 Enhancements}} \\[0.3em]
        Multi-init          & 3 seeds per fold, keep best \\
        Test-time augment   & 3 augmented passes, averaged \\
        Platt calibration   & Logistic on held-out scores \\
        \bottomrule
    \end{tabular}
\end{table}
\section{Results}
\label{sec:results}
All results are reported under five-fold stratified cross-validation on 308 ADNI-1 subjects. M1 and M2 refer to the baseline and stability-enhanced configurations defined in Sections~\ref{sec:stage1} and~\ref{sec:enhanced}, respectively.
\subsection{Overall Performance}
\noindent
\vspace{-1em}
\begin{table}[H]
    \centering
    \caption{Overall pooled performance over 308 subjects under five-fold stratified cross-validation.}
    \label{tab:overall}
    \begin{tabular}{l@{\hspace{1.5cm}}c@{\hspace{1.5cm}}c}
        \toprule
        \textbf{Metric} & \textbf{M1} & \textbf{M2} \\
        \midrule
        Accuracy             & 0.67 & 0.72 \\
        Balanced Accuracy    & 0.66 & 0.71 \\
        Precision (weighted) & 0.67 & 0.71 \\
        Recall (weighted)    & 0.67 & 0.72 \\
        F1 (weighted)        & 0.67 & 0.71 \\
        Sensitivity (AD)     & 0.59 & 0.63 \\
        Specificity (CN)     & 0.73 & 0.78 \\
        AUC-ROC              & 0.75 & 0.79 \\
        \bottomrule
    \end{tabular}
\end{table}

\subsection{Per-Fold Cross-Validation}

\begin{table}[H]
    \centering
    \caption{M1 - per-fold metrics.}
    \label{tab:m1_folds}
    \small
    \setlength{\tabcolsep}{4pt}
    \begin{tabular}{lccccccc}
        \toprule
        \textbf{Fold} & \textbf{Acc} & \textbf{Bal.\ Acc} & \textbf{AUC} & \textbf{Prec.} & \textbf{F1} & \textbf{Sens.} & \textbf{Spec.} \\
        \midrule
        1 & 0.58 & 0.57 & 0.68 & 0.58 & 0.58 & 0.53 & 0.61 \\
        2 & 0.69 & 0.69 & 0.80 & 0.70 & 0.69 & 0.75 & 0.64 \\
        3 & 0.70 & 0.70 & 0.80 & 0.70 & 0.70 & 0.62 & 0.77 \\
        4 & 0.72 & 0.71 & 0.73 & 0.72 & 0.72 & 0.70 & 0.73 \\
        5 & 0.65 & 0.62 & 0.75 & 0.68 & 0.62 & 0.33 & 0.91 \\
        \midrule
        \textbf{Mean} & \textbf{0.67} & \textbf{0.66} & \textbf{0.75} & \textbf{0.6799} & \textbf{0.66} & \textbf{0.59} & \textbf{0.73} \\
        \textbf{Std}  & 0.05 & 0.06 & 0.05 & 0.05 & 0.06 & 0.16 & 0.11 \\
        \bottomrule
    \end{tabular}
\end{table}

\begin{table}[H]
    \centering
    \caption{M2 - per-fold metrics.}
    \label{tab:m2_folds}
    \small
    \setlength{\tabcolsep}{4pt}
    \begin{tabular}{lccccccc}
        \toprule
        \textbf{Fold} & \textbf{Acc} & \textbf{Bal.\ Acc} & \textbf{AUC} & \textbf{Prec.} & \textbf{F1} & \textbf{Sens.} & \textbf{Spec.} \\
        \midrule
        1 & 0.67 & 0.67 & 0.75 & 0.67 & 0.67 & 0.60 & 0.73 \\
        2 & 0.74 & 0.74 & 0.81 & 0.74 & 0.74 & 0.78 & 0.70 \\
        3 & 0.72 & 0.70 & 0.83 & 0.73 & 0.71 & 0.51 & 0.88 \\
        4 & 0.73 & 0.71 & 0.79 & 0.74 & 0.72 & 0.55 & 0.88 \\
        5 & 0.72 & 0.71 & 0.81 & 0.72 & 0.72 & 0.70 & 0.73 \\
        \midrule
        \textbf{Mean} & \textbf{0.72} & \textbf{0.71} & \textbf{0.80} & \textbf{0.72} & \textbf{0.71} & \textbf{0.63} & \textbf{0.78} \\
        \textbf{Std}  & 0.02 & 0.02 & 0.02 & 0.03 & 0.02 & 0.10 & 0.08 \\
        \bottomrule
    \end{tabular}
\end{table}

\subsection{ROC Curves}

\begin{figure}[H]
    \centering
    \includegraphics[width=0.6\textwidth]{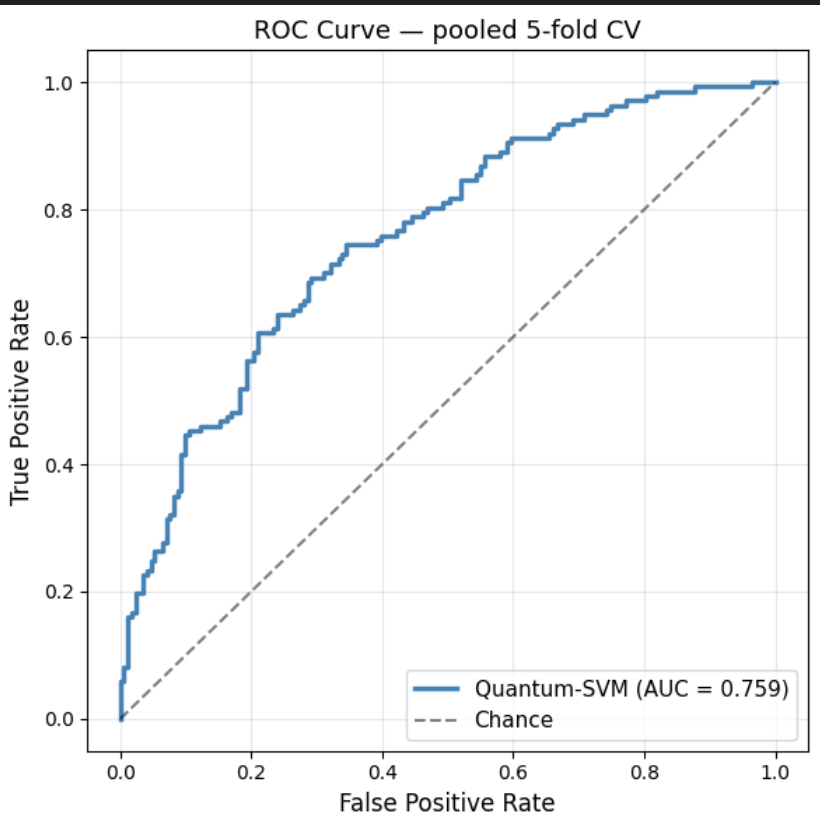}
    \caption{ROC curve for M1 (pooled 5-fold CV), AUC $=0.759$.}
    \label{fig:roc_m1}
\end{figure}

\begin{figure}[H]
    \centering
    \includegraphics[width=0.6\textwidth]{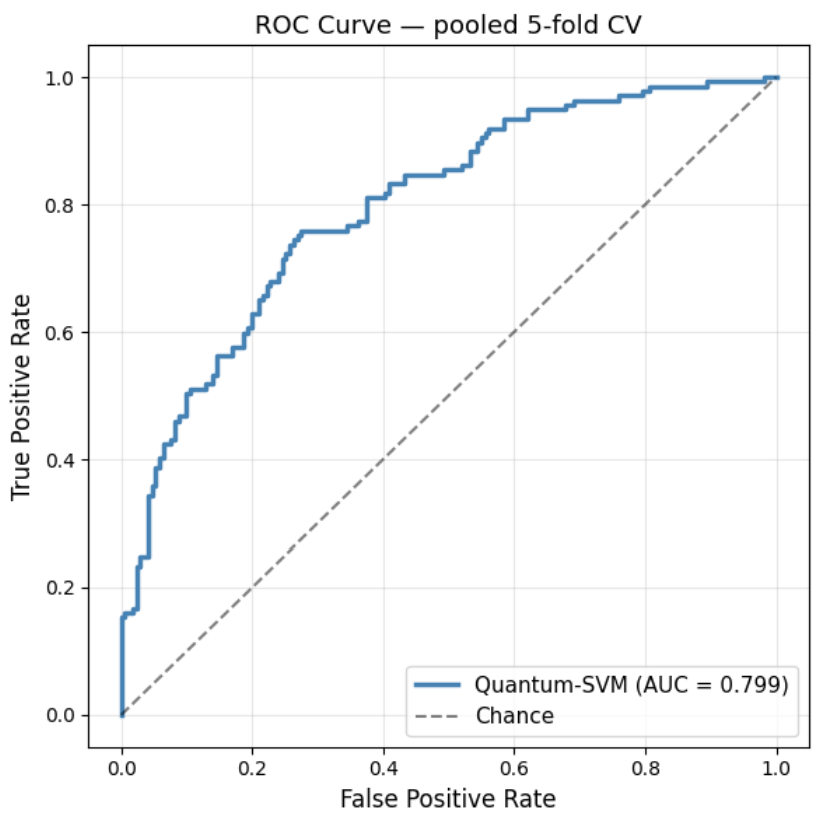}
    \caption{ROC curve for M2 (pooled 5-fold CV), AUC $=0.799$.}
    \label{fig:roc_m2}
\end{figure}

\subsection{Confusion Matrices}

\begin{table}[H]
    \centering
    \caption{Pooled confusion matrices across 308 subjects.}
    \label{tab:confusion}
    \begin{tabular}{lcccc}
        \toprule
        & \multicolumn{2}{c}{\textbf{M1(Baseline)}} & \multicolumn{2}{c}{\textbf{M2(Stability variant)}} \\
        \cmidrule(lr){2-3} \cmidrule(lr){4-5}
        & Pred CN & Pred AD & Pred CN & Pred AD \\
        \midrule
        \textbf{Actual CN} & 126 & 45 & 135 & 36 \\
        \textbf{Actual AD} & 56  & 81 & 50  & 87 \\
        \bottomrule
    \end{tabular}
\end{table}
\subsection{Interpretability using Grad-CAM}

\begin{figure}[H]
    \centering
    \includegraphics[width=0.95\textwidth]{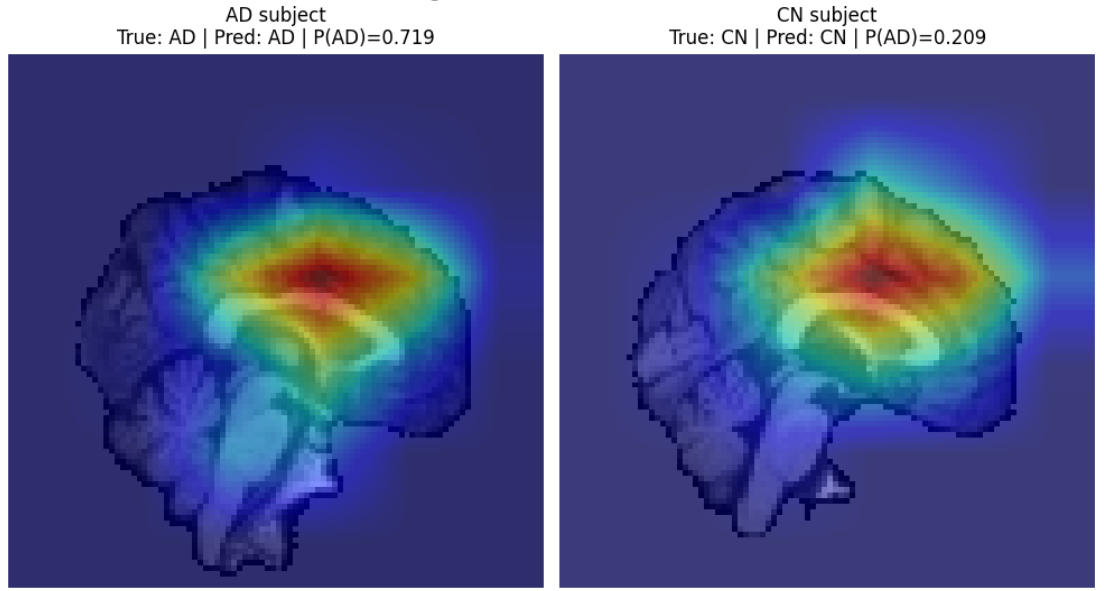}
    \caption{Grad-CAM activation maps for M1. The left represents AD subject correctly predicted as AD, $P(\text{AD})=0.719$. The right represents CN subject correctly predicted as CN, $P(\text{AD})=0.209$.}
    \label{fig:gradcam_m1}
\end{figure}

\begin{figure}[H]
    \centering
    \includegraphics[width=0.85\textwidth]{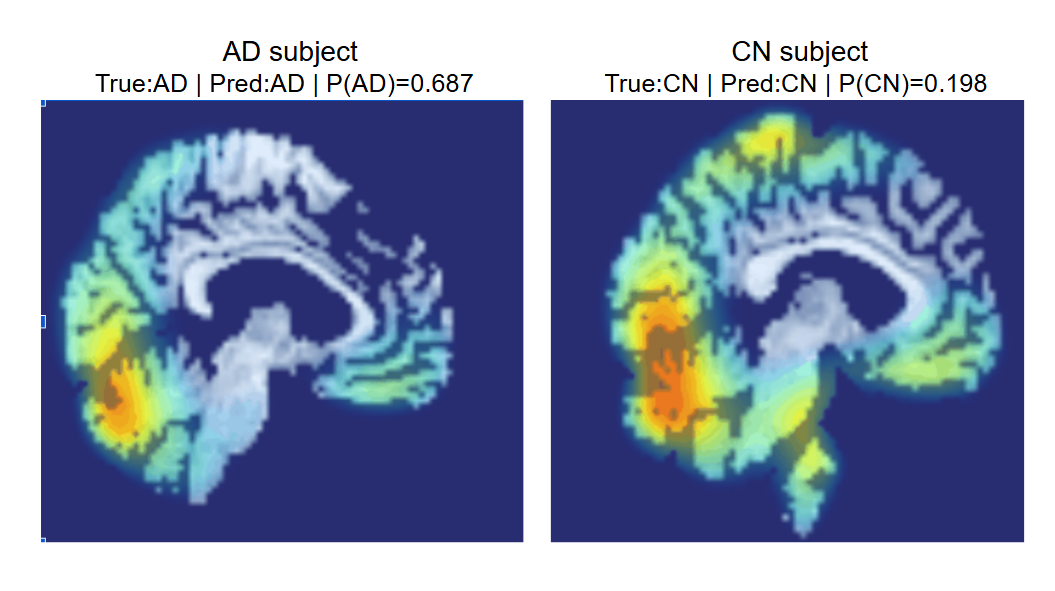}
    \caption{Grad-CAM activation maps for M2. The left represents AD subject correctly predicted as AD, $P(\text{AD})=0.687$. The right represents CN subject correctly predicted as CN, $P(\text{AD})=0.198$.}
    \label{fig:gradcam_m2}
\end{figure}

This is to verify that the classifier's predictions are driven by anatomically meaningful regions rather than noise or scanner artifacts, we apply 3D Grad-CAM to the encoder's final convolutional layer. The resulting heatmaps highlight which parts of the brain contributed most to each prediction. Figures~\ref{fig:gradcam_m1} and~\ref{fig:gradcam_m2} show representative AD and CN subjects under M1 and M2 respectively, with warmer colors indicating stronger activation.

\newpage
\section{Discussion}
\label{sec:discussion}

The results indicate that the proposed hybrid pipeline successfully extracts a disease-aware representation from 3D structural MRI for quantum-kernel-based AD classification. Both configurations perform above chance on a challenging few-hundred-subject task, with M2 improving over M1 primarily through variance reduction rather than a fundamental change in what the model learns the cross-fold standard deviation of accuracy halves from 0.057 to 0.026, and AUC from 0.051 to 0.028. Since all four stability enhancements act exclusively on the classical side, this indicates that much of M1's instability stems from training randomness on limited data rather than a limitation of the quantum kernel. The off-diagonal kernel statistics (mean pairwise overlap of 0.04--0.05) further confirm that the six-qubit ZZ kernel remained informative and did not collapse under exponential concentration~\cite{caro2022generalization,thanasilp2024exponential,schnabel2025quantum}.

A consistent pattern across both configurations is the gap between specificity and sensitivity (0.737/0.591 in M1; 0.789/0.635 in M2),the model is more reliable at confirming cognitively normal brains than at catching Alzheimer's. Even in M2, 50 of 137 AD subjects are missed against only 36 of 171 CN subjects wrongly flagged. This asymmetry likely reflects both the mild cohort imbalance toward CN and genuine biological overlap, since early or atypical AD brains are structurally similar to healthy aging brains and their latent codes naturally fall near the decision boundary. For a screening deployment, the per-fold decision threshold offers a direct lever for trading specificity against sensitivity.

The Grad-CAM analysis~\cite{selvaraju2020gradcam} provides anatomical validation of the learned representation. In both configurations, the classifier attends primarily to the medial-temporal lobe, particularly the hippocampus and parahippocampal regions and extends across the temporal and parietal cortices. These are precisely the regions where AD causes the earliest and most pronounced structural damage~\cite{frisoni2010clinical}: the hippocampus is among the first structures to atrophy, and the broader cortical thinning reflects later-stage spread consistent with established neuropathological staging. The fact that the supervised $\beta$-VAE learned to focus on these clinically established regions without any anatomical prior is strong evidence that the latent code captures genuine disease structure rather than dataset artifacts. The consistent probability separation across folds ($P(\text{AD})$ of 0.687-0.719 for AD versus 0.198-0.209 for CN) confirms this grounding is stable.

To our knowledge, this is the first evaluation of a quantum kernel on supervised, end-to-end-learned autoencoder representations for volumetric AD classification. Prior hybrid approaches operated on pre-extracted or handcrafted features~\cite{jenber2024deep,mazher2024hybrid,cappiello2025quantum,chen2025compressedmediq,islam2025cq,barua2025hybrid} with no mechanism for the representation and the quantum classifier to cooperate. Our pipeline closes this gap by ensuring that the features entering the quantum stage are shaped by the diagnostic task from the first epoch. We deliberately stop short of claiming a quantum advantage,isolating the quantum kernel's specific contribution would require the controlled comparison outlined in Section~\ref{sec:FutureWork},but the results establish that the architecture works, the quantum kernel stays informative, and the combination produces anatomically grounded predictions on real clinical data.
\section{Limitations}
\label{sec:limitations}

The study uses 308 subjects from a single source (ADNI-1) with no external cohort validation, so generalization across scanners, protocols, and populations remains untested. All quantum operations run on PennyLane's noiseless \texttt{lightning.qubit} simulator; real NISQ hardware would introduce gate noise, decoherence, and finite-shot sampling error, likely degrading performance. We do not compare the quantum kernel against a classical RBF or linear SVM on the same PLS-reduced features, so the quantum component's specific contribution cannot yet be isolated. The pipeline compresses each volume from 884,736 voxels to a 64-dimensional latent and then to just six PLS components. While necessary for the six-qubit circuit, this aggressive reduction inevitably discards information. Both configurations are more specific than sensitive (0.789/0.635 in M2), meaning roughly a third of AD cases are still missed, and the model is not yet suitable for clinical decision-making. The six-qubit register and single-repetition circuit are deliberately chosen to avoid exponential concentration~\cite{caro2022generalization,thanasilp2024exponential}, but this conservative design limits the representational capacity of the quantum stage and prevents scaling to richer feature sets without concentration-aware kernel designs.

\section{Conclusion and Future Work}
\label{sec:FutureWork}

This work demonstrated that coupling a supervised, disease-aware 
autoencoder with a quantum kernel classifier is a viable approach 
to Alzheimer's classification from 3D structural MRI, even on a 
cohort of just 308 subjects. The key finding is not the headline 
accuracy alone, which is 67.2\% for the baseline, 72.1\% for the 
stability-enhanced variant, but that the quantum kernel remained 
informative throughout training, that the stability enhancements 
halved cross-fold variance without altering the quantum 
architecture, and that Grad-CAM independently confirmed the 
model attends to medial-temporal and cortical regions consistent 
with established AD pathology. 

The most immediate next step is a controlled head-to-head comparison by running classical RBF and linear SVMs on the same PLS-reduced features to isolate the quantum kernel's specific contribution. Beyond this, external-cohort validation on AIBL or OASIS would test generalization, and evaluation on real NISQ hardware would reveal how gate noise and finite-shot sampling affect the kernel in practice. On the design side, trainable quantum embedding kernels could extract more signal from the same qubit budget, while concentration-aware designs would enable scaling beyond six qubits. Finally, extending the pipeline from AD-versus-CN to MCI detection and MCI-to-AD progression prediction would address a more clinically valuable question where early stratification has direct impact on patient care.
\newpage
\section*{Supplementary Material}

This section provides a self-contained introduction to the quantum computing concepts used in Stage 2 of the proposed pipeline. It is intended for readers with a classical machine learning background who may be encountering quantum circuits for the first time.

\subsection*{Qubits and State Representation}

A classical bit is either 0 or 1. A qubit, the quantum analog of a bit, can exist in a superposition of both states simultaneously. The state of a single qubit is written as
\begin{equation}
    |\psi\rangle = \alpha|0\rangle + \beta|1\rangle,
    \label{eq:qubit}
\end{equation}
where $\alpha$ and $\beta$ are complex numbers called amplitudes. These amplitudes determine how much the qubit leans toward each state. When the qubit is measured, it collapses to $|0\rangle$ with probability $|\alpha|^2$ and to $|1\rangle$ with probability $|\beta|^2$, subject to the normalization constraint $|\alpha|^2 + |\beta|^2 = 1$. In vector form, $|0\rangle = \begin{bsmallmatrix}1\\0\end{bsmallmatrix}$ and $|1\rangle = \begin{bsmallmatrix}0\\1\end{bsmallmatrix}$, so the qubit state is simply a unit vector in $\mathbb{C}^2$.

For a system of $n$ qubits, the joint state lives in a $2^n$ dimensional space. In our pipeline, six qubits produce a state in a $2^6 = 64$ dimensional space; this exponential scaling is what gives quantum feature maps access to a much larger representation space than classical encodings of the same six input features.

\subsection*{Quantum Gates}

Quantum gates are the building blocks of quantum circuits, analogous to logic gates in classical digital circuits. Each gate is a unitary matrix that transforms the qubit state vector. Applying a gate $U$ to a state $|\psi\rangle$ produces a new state $U|\psi\rangle$, just as multiplying a matrix by a vector produces a new vector in linear algebra. Unlike classical logic gates, quantum gates are always reversible and every gate has an inverse that undoes its effect.

\subsection*{Single-Qubit Gates}

Table~\ref{tab:single_gates} lists the single-qubit gates relevant to this work along with their matrix representations and functions.

\begin{table}[H]
    \centering
    \caption{Single-qubit gates used in the ZZ feature map.}
    \label{tab:single_gates}
    \begin{tabular}{p{2.8cm} c p{5.5cm}}
        \toprule
        \textbf{Gate} & \textbf{Matrix} & \textbf{What it does} \\
        \midrule
        Pauli-X (NOT) &
        $\begin{bmatrix} 0 & 1 \\ 1 & 0 \end{bmatrix}$ &
        Flips $|0\rangle$ to $|1\rangle$ and vice versa. The quantum equivalent of a classical NOT gate. \\[1.2em]
        \midrule

        Pauli-Z &
        $\begin{bmatrix} 1 & 0 \\ 0 & -1 \end{bmatrix}$ &
        Leaves $|0\rangle$ unchanged but flips the sign of $|1\rangle$. Changes the phase without affecting measurement probabilities. \\[1.2em]
        \midrule

        Hadamard (H) &
        $\dfrac{1}{\sqrt{2}}\begin{bmatrix} 1 & 1 \\ 1 & -1 \end{bmatrix}$ &
        Creates an equal superposition of $|0\rangle$ and $|1\rangle$. Transforms a qubit from a definite state into one where both outcomes are equally likely. \\[1.2em]
        \midrule

        $R_Z(\theta)$ &
        $\begin{bmatrix} e^{-i\theta/2} & 0 \\ 0 & e^{i\theta/2} \end{bmatrix}$ &
        Rotates the qubit around the Z-axis by angle $\theta$. Changes the relative phase between $|0\rangle$ and $|1\rangle$ without affecting their measurement probabilities. In our pipeline, $\theta$ is set to the rotation angle from PLS, so each input feature is encoded as a phase. \\[1.2em]
        \midrule

        $R_Y(\theta)$ &
        $\begin{bmatrix} \cos\frac{\theta}{2} & -\sin\frac{\theta}{2} \\ \sin\frac{\theta}{2} & \cos\frac{\theta}{2} \end{bmatrix}$ &
        Rotates the qubit around the Y-axis by angle $\theta$. Unlike $R_Z$, this gate shifts the measurement probabilities between $|0\rangle$ and $|1\rangle$. \\[1.2em]
        \midrule

        $R_X(\theta)$ &
        $\begin{bmatrix} \cos\frac{\theta}{2} & -i\sin\frac{\theta}{2} \\ -i\sin\frac{\theta}{2} & \cos\frac{\theta}{2} \end{bmatrix}$ &
        Rotates the qubit around the X-axis. Similar to $R_Y$ but introduces complex phase effects. \\
        \bottomrule
    \end{tabular}
\end{table}

\subsection*{Two-Qubit Gates}

Single-qubit gates act on individual qubits independently. To create correlations between qubits,which is essential for encoding feature interactions ,we need two-qubit entangling gates. Table~\ref{tab:two_gates} describes the key two-qubit gate used in our pipeline.

\begin{table}[H]
    \centering
    \caption{Two-qubit gate used in the ZZ feature map.}
    \label{tab:two_gates}
    \begin{tabular}{p{2.8cm} c p{5.5cm}}
        \toprule
        \textbf{Gate} & \textbf{Matrix} & \textbf{What it does} \\
        \midrule
        CNOT (Controlled-NOT) &
        $\begin{bmatrix} 1&0&0&0 \\ 0&1&0&0 \\ 0&0&0&1 \\ 0&0&1&0 \end{bmatrix}$ &
        Acts on two qubits: a control ($\bullet$) and a target ($\oplus$). If the control qubit is closer to $|1\rangle$, the target qubit is flipped; if the control is closer to $|0\rangle$, the target is left unchanged. This creates a dependency between the two qubits, allowing one feature to influence another. \\
        \bottomrule
    \end{tabular}
\end{table}

\subsection*{How These Gates Combine in the ZZ Feature Map}

The ZZ feature map used in Stage~2 of our pipeline applies these gates in a specific sequence to encode six input features into a quantum state that captures both individual feature values and their pairwise interactions. The encoding proceeds in three layers:

\textbf{Layer 1 - Hadamard.} A Hadamard gate is applied to each of the six qubits, transforming every qubit from the definite state $|0\rangle$ into an equal superposition of $|0\rangle$ and $|1\rangle$. At this point, each qubit is equally uncertain about both outcomes, and the system is ready to receive data.

\textbf{Layer 2 - $R_Z$ phase encoding.} An $R_Z(\varphi_i)$ rotation is applied to each qubit, where $\varphi_i$ is the rotation angle for the $i$-th input feature (produced by PLS in Stage-1). This writes each feature into the phase of its corresponding qubit. The measurement probabilities remain unchanged , what changes is the internal phase relationship, which will affect how qubits interact in the next layer.

\textbf{Layer 3 -ZZ entangling.} For every pair of qubits $(i, j)$, a ZZ interaction is applied. Each ZZ interaction is decomposed into three elementary gates:
\begin{equation}
    \text{ZZ}(i,j) = \text{CNOT}(i,j) \cdot R_Z\big(2(\pi - \varphi_i)(\pi - \varphi_j)\big) \cdot \text{CNOT}(i,j).
    \label{eq:zz_decomp}
\end{equation}
The first CNOT creates a temporary dependency between qubits $i$ and $j$. The $R_Z$ rotation then imprints a phase that depends on both features simultaneously this is the step that encodes the interaction between the two features. The second CNOT undoes the dependency, leaving behind only the phase contribution. Before this layer, qubit $i$ only knew about feature $i$. After it, qubit $i$ carries information about how feature $i$ relates to every other feature. With six qubits, there are $\binom{6}{2} = 15$ such pairs, and all 15 are connected.

The result is a quantum state $|\psi(x)\rangle$ in a 64-dimensional space that encodes both individual features and all pairwise interactions providing a richer representation than the original six numbers, obtained without adding any trainable parameters.
\printbibliography
\end{document}